\documentclass[runningheads]{llncs}
\usepackage[T1]{fontenc}
\usepackage{graphicx}

\usepackage{wrapfig}
\usepackage{amsmath}
\usepackage{xfrac}
\usepackage{algorithm}
\usepackage{algpseudocode}
\usepackage{subcaption}
\usepackage{amssymb}
\usepackage{booktabs}

\usepackage{tikz}
\usetikzlibrary{shapes,arrows}

\usepackage{subcaption}
\newcommand{\mv}{MultiVeStA}
\newcommand{\codefont}[1]{{\small\texttt{#1}}}
\newcommand{\cf}[1]{\codefont{#1}}

\usepackage{listings}
\definecolor{keywordsColor}{RGB}{0,0,200}

\AtBeginDocument{}

\begin{document}
	\synctex=1
	
%
\title{Towards Agentic Agent-based Models: Feasibility, Performance, and Statistical Model Checking}
%
\titlerunning{Towards Agentic Agent-based Models}
%
\author{
  Stefano Blando\inst{1,5}\orcidID{0009-0007-0523-6855} \and
  Emanuele Guerrazzi\inst{1}\orcidID{0000-0002-6545-2089} \and
  Riccardo Porcedda\inst{1,5}\orcidID{0000-0001-7360-4401}\and
  Giuseppe Squillace \inst{2} \and
  Max Tschaikowski \inst{3}\orcidID{0000-0002-6186-8669)} \and
  Andrea Vandin\inst{1,4}\orcidID{0000-0002-2606-7241}
}

\authorrunning{Blando, Guerrazzi, Porcedda, Squillace, Tschaikowski, Vandin}
%
%
\institute{
     Sant'Anna School of Advanced Studies Pisa, Italy 
\and CentraleSupélec, Université Paris-Saclay, France
\and Sapienza University of Rome, Italy
\and DTU Technical University of Denmark, Denmark
\and University of Pisa, Italy\\
\email{andrea.vandin@santannapisa.it}
}
\maketitle              
\begin{abstract}
	Agent-based models (ABMs) rely on simple, explicit and reproducible rules for individual decision making, while complex collective behavior emerges from interactions among agents. Recent advances in large language models (LLMs) make it tempting to replace, enrich, or perturb these rules with LLM-based agentic capabilities. However, this raises a methodological question: how does introducing LLM-driven decisions affect the reliability, computational cost, and behavior of ABM simulations? We investigate this for Mesa ABM models, a popular Python library for ABMs, analyzed by statistical model checking. Building on Mesa's integration with the statistical model checker MultiVeStA, we extend the classical Schelling segregation model with a hybrid population: ordinary agents classify neighbors using the standard symbolic rule, while one agent delegates this task to an LLM through tool calls. The LLM-enabled agent receives natural-language descriptions of neighboring agents and invokes tools that increment counters of similar/different neighbors; these counters determine its happiness according to the original Schelling dynamics. This provides a minimal but controlled setting where the semantic, operational, and computational behavior of LLM-based decisions can be studied inside an otherwise standard ABM. 
	We report preliminary experiments with locally served LLMs of different sizes, showing that smaller models may fail simple semantic classification experiments or become operationally unusable during repeated tool-call generation, while larger tested models pass these preliminary checks. We discuss how statistical model checking can estimate classical ABM observables and quantify the impact of introducing agentic LLM components into simulation models.
	
\end{abstract}

\section{Introduction}
\label{intro}
%

Agent-based Models (ABMs) are characterized by interactions of autonomous and
possibly heterogeneous agents, often mediated by an environment or by spatial
constraints. Although the behavior of each agent is usually specified by simple
local rules, the repeated interaction of many agents may give rise to complex
emergent behaviors that are not explicitly encoded in the model. This modeling
paradigm has been used in several disciplines, including ecology~\cite{grimm2013},
health care~\cite{eff12}, geography~\cite{brown2005}, and medicine~\cite{An2009}.
ABMs are also widely used in the social sciences
(see, e.g.,~\cite{pangallonature2024,macy2002,tesfatsion2006handbook,gatti2020rising,poledna2020economic,FagioloRoventini2012,fagiolo2017macroeconomic,dosi2019more,DBLP:journals/corr/abs-2509-10977,DOSI20101748}),
where they provide a simulation-based way to study phenomena that are too
complex to be treated analytically.
It is therefore important to provide modeling and methodological extendions for ABMs.

A key feature of many ABMs is that individual behavior is specified by explicit
and reproducible rules. This makes the model transparent: one can inspect how
agents take decisions, reproduce simulations under controlled conditions, and
study how local rules affect global outcomes.
In other disciplines, ABMs are named Multi-Agent Systems (MAS) or Collective Adaptive Systems (CAS), see, e.g., \cite{DBLP:journals/scp/NicolaSI20,roccoOmarBirds2,DBLP:conf/isola/NicolaSIV22,10.1007/978-3-031-73709-1_13,DBLP:conf/sfm/VandinT16,DBLP:journals/aamas/BaiardiBCP26}.

Recent advances in large language
models (LLMs),  make it increasingly tempting to enrich ABM agents with
more flexible, language-mediated, and agentic capabilities. 
%
Instead of hard-coding all
decisions symbolically, one may delegate part of an agent's local reasoning
to an LLM, possibly allowing the agent to interpret natural-language information,
use tools, or adapt its behavior in a less rigid way.
Surveys on
LLM-empowered ABMs explicitly frame this as an
emerging research direction, emphasizing that LLM agents can enhance perception,
reasoning, decision-making, adaptation, and heterogeneous role-playing in
multi-agent simulations \cite{gao2024large}.

This possibility is appealing, but it also raises methodological concerns. LLMs
are not standard symbolic rules: they may introduce semantic errors, stochastic
variability, prompt sensitivity, tool-call failures, and non-negligible runtime
overhead.
These concerns are consistent with the broader
challenges identified in \cite{gao2024large}.
Moreover, when an LLM is embedded inside an ABM simulation loop, its
failures may affect not only the behavior of one agent, but also the emergent
behavior of the whole system. This suggests that LLM-enabled ABM agents should
not be treated as transparent replacements for symbolic rules. Rather, their
impact should be studied explicitly and quantitatively.

Statistical model checking (SMC) provides a natural framework for this task.
SMC is a family of simulation-based analysis techniques from computer science
that automates simulation experiments and associates quantitative estimates with
statistical guarantees~\cite{LLTYSG19,Agha18}. In particular, black-box SMC only
requires that the model can be simulated probabilistically and that observations
can be evaluated on simulation states~\cite{sen2004statistical,younes2005probabilistic}.
This makes it suitable for analyzing existing simulation models without
translating them into a different formalism. MultiVeStA
\cite{sebastio2013multivesta,DBLP:conf/ifm/GilmoreRV17,VANDIN2022104458} is a
black-box SMC tool that can be integrated with external simulators. In previous
work \cite{DBLP:conf/vecos/Vandin24}, MultiVeStA was integrated with Mesa, a Python framework for ABM modeling
and simulation popular in the social sciences~\cite{python-mesa-2020}, making
SMC available for Mesa models.

In this paper, we build on this integration to study the effect of introducing
LLM-based agentic capabilities in Mesa. We focus on the classical and  influential
Schelling segregation model~\cite{doi:10.1080/0022250X.1971.9989794}.
In the standard model, agents belong to one of two groups and they move according to a rule based on the counting of how many similar neighbours they have.
We construct a minimal LLM-extended variant in which the global Schelling dynamics is left
unchanged, but one distinguished agent delegates the classification of neighbours' similarity to an LLM.

%
This setting is deliberately simple. Its purpose is not to claim that the
Schelling model requires LLM-based reasoning, nor to cover all possible forms of
agentic ABMs. Rather, it provides a controlled benchmark in which the
consequences of replacing a symbolic local decision rule with an LLM-based one
can be isolated. This allows us to study two related questions. First, can local
LLMs of different sizes reliably perform the semantic classification and
tool-call generation required by the agent? Second, when an LLM-enabled agent is
embedded in the full simulation, does it measurably affect the transient
behavior of the ABM, and at what computational cost?

\paragraph{Contribution.}
The contribution of this paper is threefold:
(i) we extend a Mesa implementation of the Schelling segregation model with an
LLM-enabled agent using tool calls; 
(ii) we study the correctness, robustness, and runtime behavior of local LLMs of
different sizes when used for the agent's local classification task;
(iii) we compare the original and LLM-extended models using statistical model
checking with MultiVeStA.

Our preliminary robustness experiments show that small local LLMs may fail even
simple semantic classification tasks, may generate the wrong tool calls, or may
become operationally unusable during multi-neighbor tool-call generation. Larger
tested variants pass these unit-style checks, but introduce a measurable runtime
overhead. In the full ABM analysis, we use MultiVeStA to estimate the transient
evolution of the ratio of happy agents in the original and LLM-extended models,
and to statistically compare the resulting trajectories.
We find that the LLM-extension does not change the model dynamics, while, as expected, it introduces very significant runtime overhead.

\emph{Synopsis.}
The paper is structured as follows. Section~\ref{sec:background} recalls the
background on Mesa, statistical model checking, MultiVeStA, and their previous
integration. Section~\ref{sec:agentic-schelling} presents the LLM-extended
Schelling model and the tool-calling mechanism used to implement the agentic
decision procedure. Section~\ref{sec:preliminary-checks} reports preliminary
correctness and robustness experiments for local Qwen3.5 models of different
sizes. Section~\ref{sec:smc-agentic} compares the original and LLM-extended
Schelling models using statistical model checking. Section~\ref{sec:concl}
concludes the paper.

%
%
%

\section{Background: Statistical model checking of Mesa models}
\label{sec:background}

This section provides the necessary background material. 
We first introduce agent-based models and Mesa, then briefly discuss statistical model
checking and MultiVeStA, and finally summarize the previous integration of Mesa
and MultiVeStA that we reuse in this work.

\subsection{Agent-Based Models and Mesa}

Agent-based models (ABMs) describe systems composed of autonomous agents
interacting with each other and, often, with an explicit environment \cite{dosi2019more}. Each agent
is typically equipped with a local state and a set of rules determining how it
acts, how it interacts with other agents, and how these interactions affect its state. Although these rules are
usually simple and specified at the level of individual agents, their repeated
interaction may give rise to complex emergent behavior that is not explicitly
encoded in the model.

ABMs are commonly used to describe systems whose complexity makes analytical treatment difficult, thus requiring simulation-based analyses. Examples include
models of flocking \cite{reynolds1987flocks}, segregation \cite{doi:10.1080/0022250X.1971.9989794}, markets \cite{bottazzigiachini2019b}, economies \cite{FagioloRoventini2012}, innovation dynamics \cite{EPTCS443.2}, epidemics \cite{hunter2017}, mobility \cite{MAGGI201650}, and many other
social or natural phenomena. A simulation of an ABM proceeds by repeatedly
activating agents, updating their state, and observing the resulting evolution of
the system. 

Mesa \cite{DBLP:conf/vecos/Vandin24} is a Python framework for the development and simulation of ABMs. A Mesa
model typically defines a class for the global model, classes for the different types of agents, and
a representation of the space or environment in which agents interact. The model
class stores global parameters and simulation state, while agents implement the
local behavior executed at each simulation step. This modular structure makes
Mesa convenient for rapidly implementing ABMs and for experimenting with
different model variants.

In this paper, we use the Mesa implementation of the popular Schelling segregation model
\cite{doi:10.1080/0022250X.1971.9989794}\footnote{As available at
	\url{https://mesa.readthedocs.io/stable/examples/basic/schelling.html}.}. We will also propose an
LLM-extended variant, presented in later sections. The use of Mesa is important
for our purposes because the LLM-enabled agent can be introduced by modifying
only the local decision procedure of one agent, while leaving the rest of the
model dynamics unchanged. This makes Mesa suitable for implementing the
controlled model variant studied below.

\subsection{Statistical Model Checking and MultiVeStA}

The analysis of ABMs is usually simulation-based: one executes the model many
times and studies the values of selected observables \cite{VANDIN2022104458}. However, the design of
simulation experiments may strongly affect the reliability of the conclusions
\cite{VANDIN2022104458,Secchi2017,EPTCS443.2}. Too few simulations,
arbitrary stopping criteria, or informal comparisons between runs may lead to
statistically weak or non-reproducible results.

Statistical model checking (SMC) \cite{LLTYSG19,Agha18} provides an automated simulation-based
approach to the quantitative analysis of stochastic systems. In black-box SMC
\cite{sen2004statistical,younes2005probabilistic}, the model under analysis is treated as a simulator: the analysis engine
does not need to inspect its internal structure, but only to run simulations and
evaluate observations on simulation states. This makes black-box SMC
particularly suitable for existing simulators and modeling frameworks, including
ABM frameworks such as Mesa \cite{DBLP:conf/vecos/Vandin24} or NetLogo
\cite{DBLP:journals/corr/abs-2509-10977}.
For example, it has been applied to scenarios as diverse as lending pools in
decentralized finance~\cite{DBLP:conf/isola/BartolettiCJLMV22}, public
transportation systems~\cite{gilmore2014analysis,ciancia2016tool},
highly-configurable systems~\cite{DBLP:journals/tse/BeekLLV20,DBLP:conf/fm/VandinBLL18},
business process modeling~\cite{DBLP:journals/jss/CorradiniFPRTV21},
security threat modeling~\cite{ter2021quantitative,10.1007/978-3-031-25383-6_18},
crowd steering scenarios~\cite{DBLP:conf/hpcs/PianiniSV14}, adaptive robotic
systems~\cite{DBLP:conf/birthday/BelznerNVW14,10.1007/978-3-031-73709-1_13,DBLP:journals/scp/BruniCGLV15},
and CAS in general~\cite{DBLP:conf/wsc/GalpinGLV18}.

MultiVeStA \cite{sebastio2013multivesta,DBLP:conf/ifm/GilmoreRV17,VANDIN2022104458}
is a statistical model checker that can be integrated with external simulators.
Given a property of interest, MultiVeStA executes a number of independent
simulations determined by the required statistical accuracy, and estimates the
expected value of the corresponding observable together with a confidence
interval. In this way, quantitative statements about the model are accompanied
by explicit statistical guarantees.

\mv{} supports two main families of analyses: \emph{transient} analysis, where
properties are studied at given points in time, and \emph{steady-state} analysis,
where properties are studied in the long run, when the system reaches a
statistical equilibrium \cite{VANDIN2022104458}. In this paper, we focus on the
former. This allows us to estimate, for instance, the expected ratio of agents
satisfying a given criterion, such as being happy, at steps $1,2,\ldots,T$,
obtaining a trajectory with confidence intervals. For transient analysis,
MultiVeStA also supports statistical comparison of different model
parameterizations or variants. This is useful here because we are interested in
comparing the original Schelling model with a variant in which one agent
delegates part of its local decision process to an LLM.

\subsection{Previous Integration of Mesa and MultiVeStA}

In previous work \cite{DBLP:conf/vecos/Vandin24}, we integrated Mesa with MultiVeStA in order to make
statistical model checking directly available for ABMs written in Mesa. The
integration follows the black-box philosophy of MultiVeStA: the Mesa model is
not translated to another formalism, but is executed by its own simulator and
queried through a small interface.

At a high level, the integration requires the Mesa model to provide three basic
operations. First, the simulator must be \texttt{reset} before each new simulation,
setting a seed provided by MultiVeStA to ensure independent and reproducible
runs. Second, it must expose an operation to execute one simulation
step, denoted by \texttt{next}. Third, the model must provide an evaluation
function, denoted by \texttt{eval}, that returns the value of a requested observable
in the current simulation state. These operations are sufficient for MultiVeStA
to orchestrate multiple simulations and compute statistically reliable estimates
of the desired properties.

The previous integration was validated on two classical ABMs: 
the Boids flocking model \cite{reynolds1987flocks} and the Schelling segregation model \cite{doi:10.1080/0022250X.1971.9989794}. The Boids model was used to illustrate transient analysis and counterfactual
comparison, while the Schelling one to illustrate steady-state
analysis and ergodicity diagnostics.

After that integration, Mesa underwent major updates, evolving from version 2
to version 3, which required updating the integration. In doing so, the
integration was made even less intrusive. 
The updated architecture introduces an additional
intermediate class, \texttt{ModelContainer}, which simplifies the integration of
new Mesa models. In the previous integration, resetting the simulator before a
new simulation did not allow the creation of a fresh \texttt{Model} instance.
Consequently, the modeler could be required to identify and clean all relevant
internal data structures inside the \texttt{reset} method. The additional
abstraction layer introduced by \texttt{ModelContainer} removes this limitation:
a new Mesa model can now be created when starting a new simulation.\footnote{The updated
	integration infrastructure is available at
	\url{https://github.com/andrea-vandin/MultiVeStA/wiki/Integration-with-Mesa}}

In the present paper, we use the same integration perspective, but with a
different objective. Our goal is not to validate the Mesa--MultiVeStA
integration again, but to analyze the original and LLM-enriched versions of the
Schelling segregation model, and to statistically compare their outcomes. To
this end, we perform transient analysis of each model version, estimating the
expected ratio of happy agents within the first $T$ steps, and comparing the
resulting trajectories to determine whether they differ from a statistical
perspective.


\section{A Schelling Model with Agentic LLM Capabilities}
\label{sec:agentic-schelling}

\subsection{The Schelling Segregation Model}
\label{subsec:schelling-model}

We consider the classical Schelling segregation model~\cite{doi:10.1080/0022250X.1971.9989794}, one of the most well-known examples of ABMs in the social sciences. The model has been encoded in several ABM frameworks, including Mesa.
It describes a population of agents divided into two groups, commonly represented by two colors, \emph{red} and \emph{blue}. These groups represent two types of similar individuals. The model is famous for showing how even a modest local preference for living close to agents of the same group can lead to global segregation patterns.

Space is abstracted as a grid, with agents occupying cell locations. Each cell can contain at most one agent. At each simulation step, an agent observes the agents located in its neighborhood and determines whether it is \emph{happy}. In the classical formulation, happiness depends on the number or fraction of neighboring agents belonging to the same group. If the agent has enough similar neighbors, it remains happy; otherwise, it becomes unhappy and moves to an empty cell. Repeated application of this local rule may lead the system to configurations in which agents of the same group form spatial clusters.

The model's behavior is controlled by a small number of parameters: the \textbf{size} of the grid; the \textbf{homophily} threshold, determining how many, or what fraction of, neighbors 
must be similar for an agent to be happy; the \textbf{density} factor, determining the probability that a cell is occupied during initialization, and thus the number of agents; and the \textbf{minority percentage}, determining the probability that a newly created agent belongs to the minority group.
\footnote{In our experiments, we use the following parametrization: 
\textbf{size}: $20\times20$, 
\textbf{homophily}: $0.7$,
\textbf{density}: $0.8$, 
\textbf{minority percentage}: $0.2$.}
In the Mesa implementation used in this paper, each agent has a symbolic type representing its group. Ordinary agents use this 
type 
to classify their neighbors: a neighbor is similar if and only if its type coincides with the type of the current agent. The result of this classification is then used to compute the agent's happiness.

Listing~\ref{lst:schelling-standard-agent} reports the relevant part of the base agent implementation. Method \cf{count\_similar\_neighbors} implements the local classification rule by iterating over the current neighbors and counting those whose field \cf{type} matches the type of the current agent. The method \cf{assign\_state} then computes the fraction of similar neighbors and updates the Boolean field \cf{happy}. Finally, the method \cf{step} moves the agent to a randomly selected empty cell whenever it is unhappy.

\begin{lstlisting}[language=Python,float=t,
	caption={Relevant excerpt of the standard Schelling agent.},
	label={lst:schelling-standard-agent}]
class SchellingAgent(CellAgent):
    """Schelling segregation agent."""

    def count_similar_neighbors(self, neighbors):
        """Count number of similar neighbors within the specified radius."""
        similar_neighbors = len([n for n in neighbors if n.type==self.type])
        return similar_neighbors

    def assign_state(self) -> None:
        """Determine if agent is happy and move if necessary."""
        neighbors=list(self.cell.get_neighborhood(radius=self.radius).agents)

        # Count similar neighbors
        similar_neighbors = self.count_similar_neighbors(neighbors)

        # Calculate the fraction of similar neighbors
        if (valid_neighbors := len(neighbors)) > 0:
            similarity_fraction = similar_neighbors / valid_neighbors
        else:
            # If there are no neighbors, the similarity fraction is 0
            similarity_fraction = 0.0

        if similarity_fraction < self.homophily:
            self.happy = False
        else:
            self.happy = True
            self.model.happy += 1

    def step(self) -> None:
        # Move if unhappy
        if not self.happy:
            self.cell = self.model.grid.select_random_empty_cell()
\end{lstlisting}

This implementation is particularly convenient for our purposes. The classification of neighbors is localized in the method \cf{count\_similar\_neighbors}, while the rest of the Schelling dynamics only depends on the number of similar neighbors returned by that method. Therefore, we can replace this local symbolic classification rule with an LLM-based one, while leaving unchanged the computation of happiness, the movement rule, and the overall model dynamics. This makes the Schelling model a controlled setting for studying the impact of introducing agentic LLM capabilities into an otherwise standard ABM.

\subsection{Local LLM Infrastructure: Ollama and Qwen3.5}
\label{subsec:ollama-qwen}

The LLM-enabled variant of the Schelling model requires repeated interactions
between the simulator and an LLM. In this work, we rely on locally served models
rather than on remote APIs. This choice is important for two reasons. First, it
makes the experimental setup more controllable, since simulations do not depend
on the availability, latency, or possible changes of an external service. Second,
it allows us to explicitly measure the computational overhead introduced by
LLM-based decisions inside the ABM simulation loop.

We use Ollama\footnote{\url{https://ollama.com/}} as local serving infrastructure for the LLMs. Ollama provides a
simple way to download, run, and query LLMs locally, and exposes interfaces that
can be used directly from Python. In our implementation, the Mesa model invokes
the locally served model whenever the LLM-enabled agent has to classify its
neighbors. Therefore, each LLM call becomes part of the simulation dynamics and
contributes to the overall runtime of the model.

As LLM family, we consider Qwen3.5 \cite{Qwen3TechnicalReport} models distributed through Ollama. 
This  is due to the fact that Qwen3.5 is distributed in several variants, with a fine range of models sizes, including those with
\texttt{0.8b}, \texttt{2b}, \texttt{4b}, \texttt{9b}, which can be run on laptop machines.
\footnote{\url{https://ollama.com/library/qwen3.5}.} 
Such families of models are often obtained by training or adapting models at different scales, possibly using techniques such as \emph{distillation} \cite{distill}. 
This range of sizes is useful for our study because it allows us to investigate
the trade-off between computational cost and reliability of the LLM-enabled
agent. Smaller models are cheaper to execute, but may be less reliable in
semantic classification and tool-call generation. Larger models may provide more
robust behavior, but require more memory and longer execution times.



\subsection{Tool Calls for Agentic Decisions}
\label{subsec:tool-calls}

The key functionality we use is \emph{tool-calling}. In a tool-calling
interaction, the LLM does not only produce textual output: it may also request
the execution of predefined functions exposed by the Python program. The Python
runtime then executes the requested functions and uses their effects/outputs to continue the computation. In our setting, this mechanism is used to
connect the linguistic decision of the LLM to the internal state of the simulated
agent. The LLM receives a natural-language description of the current
neighborhood and produces calls to tools that update the counters of similar and
different neighbors.


In our implementation, the LLM is created once in the constructor of the
\cf{Schelling\_LLM} model class, which extends the original \cf{Schelling} model class. The LLM
is stored in an attribute of the model, \cf{llm}, and can therefore be reused by all
LLM-enabled agents. This design is useful also for future extensions with more
than one LLM-enabled agent, since all of them may share the same local LLM
instance. The LLM is initialized as follows.

\begin{lstlisting}[language=Python,float=t,
	caption={Creation of the local LLM and binding of the available tools.},
	label={lst:llm-bind-tools}]
from langchain_ollama import ChatOllama
self.llm = ChatOllama(   model=llm_model_name,   validate_model_on_init=True,   
                         disable_streaming=True, temperature=0
  ).bind_tools([increment_similar_count, increment_different_count])
\end{lstlisting}

The parameter \cf{llm\_model\_name} selects the  local model to use,
e.g. a Qwen3.5 variant discussed above or any other LLM model available in Ollama. We set the temperature to
zero in order to reduce the variability of individual LLM answers and make the
experiments more reproducible. Intuitively, temperature controls the randomness
of the decoding process: lower values make the model more likely to select
high-probability outputs. The call to \cf{bind\_tools} exposes two Python
functions to the model: one for incrementing the number of similar neighbors and
one for incrementing the number of different neighbors.

These two tools are shown in Listing~\ref{lst:schelling-tools}. They receive the
current agent as argument and update one of its counters. The return values are
not used to control the model dynamics directly; the relevant effect is the
update of the internal counters of the current agent.

\begin{lstlisting}[language=Python,float=t,
	caption={Tools exposed to the LLM-enabled Schelling agent.},
	label={lst:schelling-tools}]
from langchain.tools import tool

@tool
def increment_similar_count(current_agent):
    """Increments the number of agents similar to the current agent.

    Args:
        current_agent: the agent for which to increment the similar count """
    current_agent.similar_neighbors += 1
    return 1

@tool
def increment_different_count(current_agent):
    """Increments the number of agents different from the current agent.

    Args:
        current_agent: the agent for which to increment the different count
    """
    current_agent.different_neighbors += 1
    return 0

TOOL_MAP = {
    "increment_similar_count": increment_similar_count,
    "increment_different_count": increment_different_count,
}
\end{lstlisting}

The resulting interaction pattern is the following. First, the simulation
constructs a prompt describing the current local situation of the LLM-enabled
agent. In our case, this prompt contains the natural-language descriptions of
the neighboring agents. Second, the LLM receives the prompt and emits a sequence
of tool calls. Third, the Python runtime executes the requested tools by looking
up their names in \cf{TOOL\_MAP}. Finally, the execution of the tools updates the
fields \cf{similar\_neighbors} and \cf{different\_neighbors} of the current
agent.

This design keeps the intervention of the LLM local and explicit. The LLM is not
allowed to move agents, change the grid, or directly determine whether an agent
is happy. It can only request calls to the two available counter-update
functions. The standard Schelling dynamics then uses the resulting counter
values to compute happiness and movement, as in the original model.

\subsection{Extending Schelling Agents with Agentic LLM-Based Capabilities}
\label{subsec:agentic-schelling-extension}

We now describe how the standard Schelling agent is extended with LLM-based
agentic capabilities. The extension is deliberately minimal. The original
Schelling dynamics is preserved: agents still occupy cells in a grid, observe
their neighbors, compute whether they are happy, and move to an empty cell when
unhappy. The only modified component is the local procedure used by one agent to
classify neighbors as similar or different.

In the standard model, each agent has a symbolic type, and neighbor
classification is performed by directly comparing symbolic types. In the
LLM-extended model, each agent still has this symbolic type, but it is also
associated with a natural-language self-description. In our implementation,
agents of type 0 describe themselves with the sentence \cf{Sky is beautiful},
while agents of type 1 use the sentence \cf{Fire is beautiful}. Ordinary agents
continue to classify neighbors symbolically. The LLM-enabled agent, instead,
receives the natural-language descriptions of its neighbors and delegates the
classification task to the LLM through the tool-calling mechanism described
above.

The model class \cf{Schelling\_LLM} extends the class \cf{Schelling} from the original Schelling model, which in turn extends the \cf{Model} class from Mesa. It creates agents of type \cf{SchellingAgent\_LLM}, which extends the standard \cf{SchellingAgent}.
In the
experiments reported below, one agent is created with \cf{use\_llm=True}, while
all other agents are created with \cf{use\_llm=False}. This makes all agents
instances of the same extended class, while only one of them actually uses the
LLM during the classification of neighbors.

Listing~\ref{lst:schelling-llm-agent} reports a compact version of the
agent. The class defines a common prompt, describing the general
role of the simulated individual, and two role-specific prompts, one for the
blue team and one for the red team. The constructor combines the common and
role-specific prompts depending on the symbolic type of the agent, and assigns
the corresponding natural-language self-description.

\begin{lstlisting}[language=Python,
	caption={Compact version of the LLM-extended Schelling agent.},
	label={lst:schelling-llm-agent}]
from langchain.messages import AIMessage, HumanMessage, SystemMessage

class SchellingAgent_LLM(SchellingAgent):
    commonPrompt = """
    You are an agent in a simulation model made for a scientific research.
    You are not an assistant to a user; you are a simulated individual
    interacting with other agents.	
    The model simulates an interaction with another agent by feeding you
    a message generated by the other agent. You have to read the message
    and decide whether to consider the other agent 'similar' or 'different'.
    Depending on your opinion, you shall update the count of similar or
    different agents."""

    bluePrompt = """
    Your role is "Blue Team": you have a particular passion for blue color:
    whatever is blue, you like. You do not like other colors."""

    redPrompt = """
    Your role is "Red Team": you have a particular passion for red color:
    whatever is red, you like. You do not like other colors."""

    def __init__(self, model, cell, agent_type: int,
                 homophily: float = 0.4, radius: int = 1,
                 use_llm=True) -> None:
        super().__init__(model, cell, agent_type, homophily, radius)
        if agent_type == 0:
            self.system_prompt = self.commonPrompt + self.bluePrompt
            self.describe_yourself = "Sky is beautiful."
        else:
            self.system_prompt = self.commonPrompt + self.redPrompt
            self.describe_yourself = "Fire is beautiful."
        self.use_llm = use_llm

    def count_similar_neighbors(self, neighbors):
        if not self.use_llm:
            self.similar_neighbors=super().count_similar_neighbors(neighbors)
        else:
            self.similar_neighbors = 0
            self.different_neighbors = 0
            neighbor_descriptions = [n.describe_yourself for n in neighbors]
            n_neighbors = str(len(neighbors))
            prompt = "You have " + n_neighbors + " neighbors. "
            prompt += "This is the list of messages received by each other agent. "
            prompt += "Produce all " + n_neighbors + " tool calls needed to update "
            prompt += "the count of similar and different neighbors:\n\t"
            prompt += str(neighbor_descriptions)
            messages = [SystemMessage(content=self.system_prompt), 
                        HumanMessage(content=prompt)]
            result = self.model.invoke(messages)
            if isinstance(result, AIMessage) and result.tool_calls:
                for tool_call in result.tool_calls:
                    tool_fn = TOOL_MAP[tool_call["name"]]
                    tool_args = {'current_agent': self}
                    tool_fn.invoke(tool_args)
            return self.similar_neighbors
\end{lstlisting}

The key method is \cf{count\_similar\_neighbors}. Its interface is the same as in
the original Schelling agent, which allows the rest of the model to remain
unchanged. If \cf{use\_llm} is false, the method simply delegates to the original
symbolic implementation. If \cf{use\_llm} is true, the counters of similar and
different neighbors are first reset. The agent then collects the
natural-language descriptions of its current neighbors and constructs a prompt
asking the LLM to produce one tool call per neighbor. The returned tool calls
are executed by looking up the corresponding Python functions in \cf{TOOL\_MAP}.
Their execution updates the fields \cf{similar\_neighbors} and
\cf{different\_neighbors} of the current agent.

After the method returns, the remaining Schelling dynamics is exactly the same
as in the standard implementation described in
Listing~\ref{lst:schelling-standard-agent}. The value returned by
\cf{count\_similar\_neighbors} is used by \cf{assign\_state} to compute the
fraction of similar neighbors and determine whether the agent is happy. If the
agent is unhappy, the standard movement rule is applied. Thus, the LLM does not
control the scheduler, the grid, or the movement policy. It only replaces one
localized symbolic decision procedure with a tool-mediated, language-based one.

This design has two advantages for our analysis. First, it provides a controlled
way to study the impact of LLM-based agentic capabilities, since the original and
extended models differ only in one local decision mechanism. Second, it allows
us to separate failures of the LLM component from the rest of the ABM dynamics:
semantic misclassifications, missing tool calls, or repeated-generation failures
can be attributed to the LLM-enabled classification step rather than to changes
in the global model structure.

\section{Robustness Experiments}
\label{sec:preliminary-checks}



Before studying the impact on emergent behaviours of the LLM injection into the ABM, we must first establish whether the LLM can reliably perform the local
task it is given. In particular, we are interested in three possible outcomes:
the LLM may produce the correct tool call, it may produce an erroneous tool
call, or it may fail operationally, for instance by getting stuck during
generation.

The following subsection describes the experimental protocol. The results are
then discussed in Section~\ref{subsec:preliminary-results}.

\subsection{Experimental Protocol}
\label{subsec:preliminary-protocol}

We consider
four local LLMs from the Qwen3.5 family: 0.8b, 2b, 4b, and 9b.
%
For each model, we instantiate the LLM with the same configuration used in the
ABM experiments: local execution through Ollama, temperature set to zero, and
the two counter-update tools bound to the model.


Each experiment consists of a system prompt and a user message (Listing \ref{lst:schelling-llm-agent}). The former is
obtained by combining the common simulation prompt with one of the two
role-specific prompts. 
In the \emph{blue} condition, the LLM is instructed to
play a blue-team agent, which likes blue and does not like other colors. The \emph{red} condition is similar. 
The same user message is tested under
both conditions.

We consider four user messages of increasing semantic difficulty:
\begin{enumerate}
    \item[\textbf{M1}] \cf{I like blue. I do not like red.}
    \item[\textbf{M2}] \cf{The sky is beautiful. The fire is not}
    \item[\textbf{M3}] \cf{The sky is beautiful.}
    \item[\textbf{M4}] \cf{You have 3 neighbours. This is the list of messages received by each other agent. Produce all 3 tool calls needed to update the count of similar and different neighbors. ['Sky is beautiful.', 'Sky is beautiful.', 'Fire is beautiful.']}
\end{enumerate}
The first message
encodes the simplest experiment possible: it explicitly mentions a preference
for blue and a dislike for red. 
The second and third messages are less direct, since the model must connect
terms such as \emph{sky} and \emph{fire} to the role-specific preferences
encoded in the system prompt.

The fourth message type, \textbf{M4}, is meant to test whether increasing
the semantic difficulty of user messages, by requiring the generation of more tool calls, negatively affects reliability.
Thus, \textbf{M4} is both the most difficult robustness check and also the
closest to the ABM setting considered in the simulation: in fact, an agent updates its
state by considering multiple neighbors.
It must be highlighted that, while \textbf{M4} considers only three neighbours, in the real ABM simulation the agent can have up to eight neighbors. 

For each pair consisting of an LLM model and a user message, we perform three
iterations with the blue role and three iterations with the red role. 

Repeating the same test is useful even with temperature set to zero, because tool-call
generation may 
exhibit implementation-level variability, and because
operational failures such as non-termination are themselves relevant for the
robustness of the approach. We also measure the 
wall-clock time 
of each
iteration.

Listing~\ref{lst:preliminary-test-function} reports the core function used for our 
preliminary checks. The function builds the message sequence, invokes
the LLM, prints the generated tool calls, and executes them via
\cf{TOOL\_MAP}. In these 
experiments, executing a tool only prints the
corresponding action, allowing us to inspect whether the selected tool is the
expected one.
\begin{lstlisting}[language=Python,float=t,
caption={Core function used for preliminary tool-call checks.},
label={lst:preliminary-test-function}]
def test_interaction_and_tool_call(system_prompt, user_query, llm):
    messages = [ SystemMessage(content=system_prompt),
        	 HumanMessage(content=user_query)      ]

    print("Invoking LLM")
    ai = llm.invoke(messages)

    print("Tool calls:", getattr(ai, "tool_calls", None))
    print("Result:", ai.content)

    if isinstance(ai, AIMessage) and ai.tool_calls:
        for tc in ai.tool_calls:
            tool_fn = TOOL_MAP[tc["name"]]
            tool_args = tc.get("args", {}) or {}
            tool_fn.invoke(tool_args)
\end{lstlisting}
We classify each iteration into one of three outcomes:
\begin{itemize}
	\item \emph{correct}: the LLM generates the expected tool call or sequence of
	tool calls;
	\item \emph{erroneous}: the LLM terminates but generates an incorrect tool
	call, a tool call inconsistent with the assigned role, or an incorrect
	number of tool calls;
	\item \emph{stuck}: the LLM does not complete the generation in a usable way.
\end{itemize}
For \textbf{M1}-\textbf{M3}, the expected behavior is one tool call:
\cf{increment\_similar\_count} if the message is compatible with the role of the
current agent, and the other function 
otherwise. For \textbf{M4}, the expected behavior is a sequence of three tool calls,
one for each neighbor description. Since the first two descriptions are
\cf{Sky is beautiful.} and the last is \cf{Fire is beautiful.}, the expected calls are two
\cf{increment\_similar\_count} calls and one \cf{increment\_different\_count} for a blue-team agent, and vice versa for a red-team one.  Therefore, producing only one or two tool calls is considered erroneous, even if the generated calls
are individually of the correct type. Since the tools used in these preliminary checks do not identify the specific neighbor to which a call refers, we evaluate correctness by considering the resulting number of similar and different calls, not their order.

\subsection{Results}
\label{subsec:preliminary-results}

We now discuss the preliminary robustness and runtime experiments. 

\subsubsection{Robustness.}
Table~\ref{tab:preliminary-outcomes} summarizes the qualitative outcomes. We
write \emph{C} for correct, \emph{SE} for semantic error, \emph{AE} for arity
error, and \emph{ST} for stuck generation. Semantic and arity error means that the model
generates, respectively, a wrong tool call, or a wrong
number of calls. Errors are marked in red. 

\begin{table}[t]
\centering
\caption{Outcomes of the preliminary correctness and robustness checks. Each
entry reports the result for the blue role and for the red role.}
\label{tab:preliminary-outcomes}
\begin{tabular}
	{l cccc cccc}
\toprule
 & \multicolumn{2}{c}{\textbf{M1}} & \multicolumn{2}{c}{\textbf{M2}} & \multicolumn{2}{c}{\textbf{M3}} & \multicolumn{2}{c}{\textbf{M4}} \\
 \cmidrule(r){2-3} \cmidrule(r){4-5} \cmidrule(r){6-7} \cmidrule(r){8-9}
\multicolumn{1}{c}{\emph{Model}} & \multicolumn{1}{c}{\emph{B}} & \multicolumn{1}{c}{\emph{R}} & \multicolumn{1}{c}{\emph{B}} & \multicolumn{1}{c}{\emph{R}} & \multicolumn{1}{c}{\emph{B}} & \multicolumn{1}{c}{\emph{R}} & \multicolumn{1}{c}{\emph{B}} & \multicolumn{1}{c}{\emph{R}} \\
%
\cmidrule(r){2-3} \cmidrule(r){4-5} \cmidrule(r){6-7} \cmidrule(r){8-9}
\cf{qwen3.5:0.8b\ \ }
& \   3C & \     \ {\color{red}ST}
& \   3C & \   {\color{red}3SE}
& \   {\color{red}3SE} & \   {\color{red}3SE}
& \   3C & \   {\color{red}3AE} \\
\cf{qwen3.5:2b}
& \   3C & \   3C 
& \   3C & \   {\color{red}3SE} 
& \   3C & \   3C 
& \   {\color{red}ST} & \   \ {\color{red}ST} \\
\cf{qwen3.5:4b}
& \   3C & \   3C
& \   3C & \   3C
& \   3C & \   3C
& \   3C & \   3C \\
\cf{qwen3.5:9b}
& \   3C & \   3C
& \   3C & \   3C
& \   3C & \   3C
& \   3C & \   3C \\
\bottomrule
\end{tabular}
\end{table}

\paragraph{Responses to \textbf{M1}.}
All models except \cf{qwen3.5:0.8b} complete
all six iterations correctly. 
The smallest model succeeds for the blue role, but
gets stuck already on the first red-role iteration. 
Thus, even the most direct
semantic experiments exposes an operational fragility in the smallest model.

\paragraph{Responses to \textbf{M2}.}
Both \cf{qwen3.5:0.8b} and \cf{qwen3.5:2b} behave correctly for the blue role, but
systematically produce the wrong tool call for the red role: they call
\cf{increment\_similar\_count} instead of the other. 
In
contrast, \cf{qwen3.5:4b} and \cf{qwen3.5:9b} complete all iterations correctly.

\paragraph{Responses to \textbf{M3}.}
The model \cf{qwen3.5:0.8b} fails
systematically in both roles, selecting always the opposite tool. 
Larger models 
complete all iterations correctly.

\paragraph{Responses to \textbf{M4}.}
The model \cf{qwen3.5:0.8b} produces the correct number of
similar and different calls for the blue role, but generates only two calls for
the red role. 
The model \cf{qwen3.5:2b} gets stuck in both roles, and therefore cannot be used reliably for this setting. 
In contrast,
\cf{qwen3.5:4b} and \cf{qwen3.5:9b} generate the expected number and type of tool
calls in all iterations and for both roles.

\subsubsection{Runtime.}
Table~\ref{tab:preliminary-runtimes} reports average runtimes, in seconds, for
the completed iterations. Stuck executions are omitted. 
Experiments were run on an Apple M4 machine with 24GB of unified memory. 
These values should not be interpreted as a detailed benchmark, since they depend on the local machine
and on the specific serving infrastructure. Nevertheless, they give a useful
indication of the overhead introduced by LLM-based decisions.

\begin{table}[t]
	\centering
	\caption{Average runtime in seconds for completed preliminary checks. Stuck
		executions are omitted.}
	\label{tab:preliminary-runtimes}
	\begin{tabular}{l cccc}
		\toprule
		\multicolumn{1}{c}{\emph{Model}} 
		& \multicolumn{1}{c}{\textbf{M1}} 
		& \multicolumn{1}{c}{\textbf{M2}} 
		& \multicolumn{1}{c}{\textbf{M3}} 
		& \multicolumn{1}{c}{\textbf{M4}} \\
		\midrule
		\cf{qwen3.5:0.8b\ \ } & 6.69 & 6.30 & 5.88 & 7.49 \\
		\cf{qwen3.5:2b}       & 7.73 & 9.55 & 9.47 & {--} \\
		\cf{qwen3.5:4b}       & 5.45 & 9.88 & 5.38 & 11.66 \\
		\cf{qwen3.5:9b}       & 9.54 & 10.61 & 8.64 & 15.60 \\
		\bottomrule
	\end{tabular}
\end{table}

\subsubsection{Discussion.}
Overall, these preliminary experiments show that the LLM component cannot be treated
as a transparent replacement for a symbolic rule. The smallest model,
\cf{qwen3.5:0.8b}, is both semantically unreliable and operationally fragile.
The model \cf{qwen3.5:2b} handles some single-message experiments, but fails on one
indirect classification task and gets stuck on the multi-neighbor tool-call
task. The two larger tested variants, \cf{qwen3.5:4b} and \cf{qwen3.5:9b}, pass
all preliminary checks. These results motivate using only models that pass these
unit-style checks in the full ABM experiments.
%
The runtime analysis suggests using \cf{qwen3.5:4b}, rather than
\cf{qwen3.5:9b}, in the full simulations.

%
%
%
%
%
%
%

\section{Statistical model checking} 
\label{sec:smc-agentic}
Here we perform statistical model checking with MultiVeStA of the original and LLM-extended model. We report on transient analyses on the (expected value of the)
ratio of happy agents in all simulation steps from 1 to 60. 
Considering the relatively high runtime per LLM invocation shown for \textbf{M4} in Table \ref{tab:preliminary-runtimes}, in order to keep SMC runtime reasonable, we performed these experiments using only one LLM-extended agent, and  a more powerful machine, namely an NVIDIA DGX Sparx with 120GB of unified memory. 
MultiVeStA is parametrized such that for each estimation it computes a confidence interval of width at most 0.1 and statistical confidence of 95\%.
In order to understand whether the LLM-extension leads to statistically significant changes in the system dynamics, we also check whether the estimates are equal point-wise across the two models. We do this via t-tests, as discussed here \cite{VANDIN2022104458}.
We also report runtimes.

\begin{figure}[t]
\centering
\includegraphics[width=0.7\textwidth]{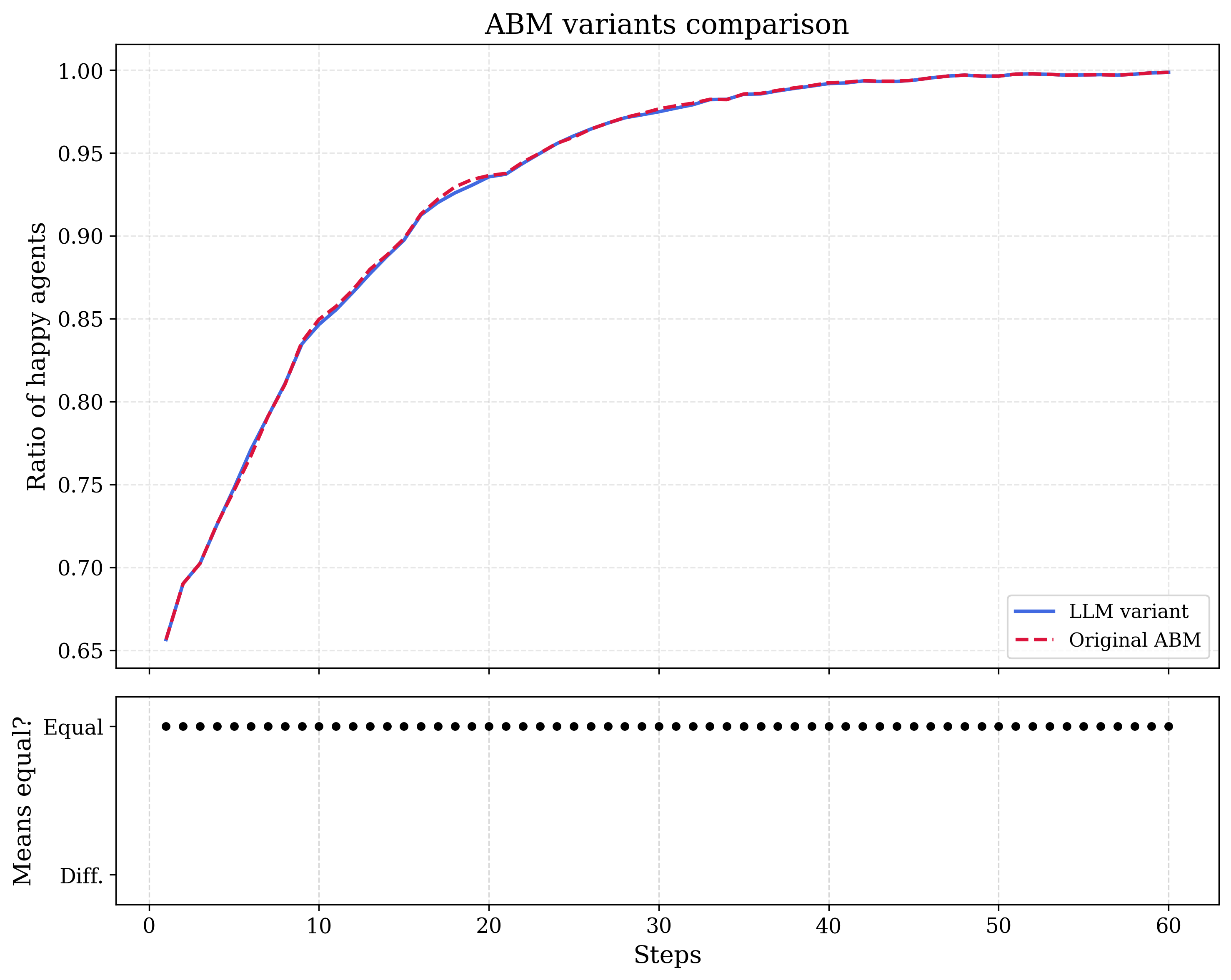}
\caption{Statistical model checking results, and t-tests\label{fig:smc}}
\end{figure}

Figure \ref{fig:smc} shows the analysis results. We can see that the two model variants produce visually-undistinguishable trajectories where the ratio of happy agents starts from about 0.65 and stabilizes at about 1.0 within 60 simulation steps.
This is confirmed by the t-tests results shown in the bottom part of the figure. 

Both SMC analyses required only 20 simulations, meaning that, with the used parametrization, the model has little variance. The runtime for the original model is just about 2 seconds, while that of the LLM-enriched model is 8917 seconds.
It is interesting to note that the LLM-enriched variant performed 61 LLM invocations (to move from step 0 to step 60) for each of the 20 simulations, that is, about 7.3 seconds per invocation. 



\section{Conclusions}\label{sec:concl}

Large language models (LLMs) make it tempting to enrich agent-based models with
language- and tool-mediated decisions. However, LLM-enabled agents are
not transparent replacements for symbolic rules: they may introduce semantic
errors, operational failures, variability, and runtime overhead. We studied
these issues by extending a Mesa implementation of the Schelling segregation
model with one LLM-enabled agent, analyzing it with statistical model
checking (SMC).

The extension was deliberately minimal: the original Schelling dynamics was left
unchanged, while one local decision procedure---the classification of neighbors
as similar or different---was replaced by an LLM-based tool-calling mechanism.
Before full ABM simulations, we performed unit-style robustness checks. Smaller
local models, \cf{qwen3.5:0.8b} and \cf{qwen3.5:2b}, exhibited semantic errors,
stuck generations, or failures in multi-neighbor tool-call generation. In
contrast, \cf{qwen3.5:4b} and \cf{qwen3.5:9b} passed all preliminary checks,
with \cf{qwen3.5:4b} offering a better runtime compromise.

We then used SMC, as implemented in MultiVeStA, to analyse the 
model variants, and to check whether the LLM-extension led to statistically significant differences in the model dynamics. The two 
model variants produce same dynamics, with significant runtime overhead due to the many LLM invocations.

This work is a first step towards a systematic analysis of agentic ABMs. Future
work will consider richer decision tasks, more LLM-enabled agents, different
ABMs, and systematic prompt and tool-interface designs. For example, here we used standard setups for the LLMs, while it could be interesting to study the impact on robustness and runtime of settings such as explicit enabling/disabling of \emph{thinking} mode. We will also investigate
process mining techniques to explain SMC results~\cite{DBLP:journals/jss/CasaluceBCLV24}, including stochastic ones \cite{DBLP:journals/is/IncertoVA25}. We will also consider the application of anomaly detection techniques \cite{frank} on the time series generated by SMC simulations. 
Overall, LLM-enabled agents should be treated as stochastic and fallible
components whose correctness, robustness, and computational cost must be
analyzed explicitly.

%
%
%
\bibliographystyle{splncs04}
\bibliography{bibitems}

\end{document}